\documentclass{article}
\usepackage{spconf,amsmath,graphicx,caption}
\usepackage{amssymb}
\usepackage[breaklinks,colorlinks]{hyperref}
\usepackage{enumerate}
\usepackage{bibspacing}
\usepackage{lipsum}
\usepackage{booktabs, multirow, multicol}
\usepackage{bbding}
\usepackage{makecell}
\newenvironment{sequation}{\begin{equation}\small}{\end{equation}}

\title{Wavelet-Decoupling Contrastive Enhancement Network for Fine-Grained Skeleton-Based Action Recognition}
\name{Haochen Chang$^{1}$ \quad Jing Chen$^{1,*}$ \quad Yilin Li$^{1}$ \quad Jixiang Chen$^{1}$ \quad Xiaofeng Zhang$^{2}$ \thanks{$^*$Corresponding Author. Email: chen74jing29@bit.edu.cn.} }

\address{$^{1}$ Beijing Engineering Research Center of Mixed Reality and Advanced Display, \\
	            School of Optics and Photonics, Beijing Institute of Technology, China \\ 
	     $^{2}$ School of Electronic Information and Electrical Engineering, Shang Hai Jiao Tong University, China}

\begin{document}
%
\maketitle

\begin{abstract}
Skeleton-based action recognition has attracted much attention, benefiting from its succinctness and robustness. However, the minimal inter-class variation in similar action sequences often leads to confusion. The inherent spatiotemporal coupling characteristics make it challenging to mine the subtle differences in joint motion trajectories, which is critical for distinguishing confusing fine-grained actions. To alleviate this problem, we propose a Wavelet-Attention Decoupling (WAD) module that utilizes discrete wavelet transform to effectively disentangle salient and subtle motion features in the time-frequency domain. Then, the decoupling attention adaptively recalibrates their temporal responses. To further amplify the discrepancies in these subtle motion features, we propose a Fine-grained Contrastive Enhancement (FCE) module to enhance attention towards trajectory features by contrastive learning. Extensive experiments are conducted on the coarse-grained dataset NTU RGB+D and the fine-grained dataset FineGYM. Our methods perform competitively compared to state-of-the-art methods and can discriminate confusing fine-grained actions well.
\end{abstract}

\section{INTRODUCTION}
\label{sec:intro}

In recent years, human action recognition has seen widespread application in various fields, such as video surveillance, VR/AR, and sports analysis. Skeleton data is more robust in handling complex backgrounds meanwhile retaining rich action information. Therefore, skeleton-based action recognition attracted much attention. Early RNN \cite{PLSTM} or CNN \cite{HCN} methods constructed skeleton data as a pseudo-image or a feature sequence, but they failed to effectively capture the structured dependencies of skeletons. Later, the GCN-based methods \cite{Body, STGCN, 2SAGCN, CTRGCN, MS-G3D,EfficientGCN-B4,SGP,InfoGCN,HD-GCN} treat human skeletons as graphs and leverage the topological structure of the skeleton to aggregate features of related joints and time series, resulting in better performances. Moreover, self-attention-based methods \cite{STTR,FOCUS,SFA} facilitate capturing long-range relationships within skeleton sequences, allowing for the extraction of global spatial-temporal features compared with GCN-based methods.

\begin{figure}[t]
	\centering
	\includegraphics[width=\linewidth]{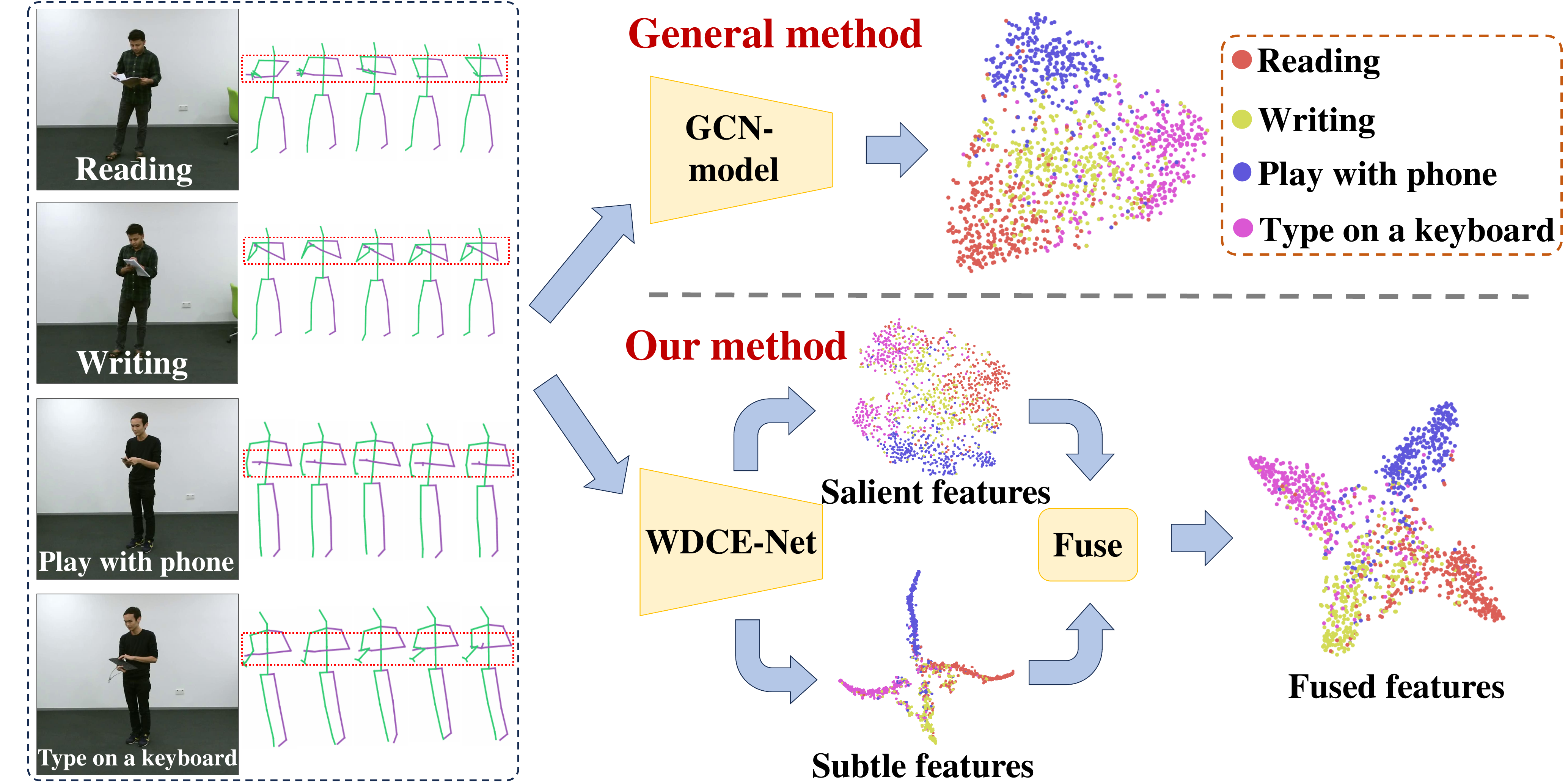}
	\vspace{-6mm}  
	\caption{The salient features of fine-grained actions frequently exhibit high similarity, but the distinctions primarily manifest in the subtle features within the red box. WDCE-Net decouples the above two features in the frequency domain and focuses on enhancing subtle features. Compared with traditional methods, our method can cluster fine-grained action features better.}
	\vspace{-6mm}  
	\label{fig1}
\end{figure} 
\begin{figure*}[htb]
	\centering
	
	\includegraphics[width=\textwidth]{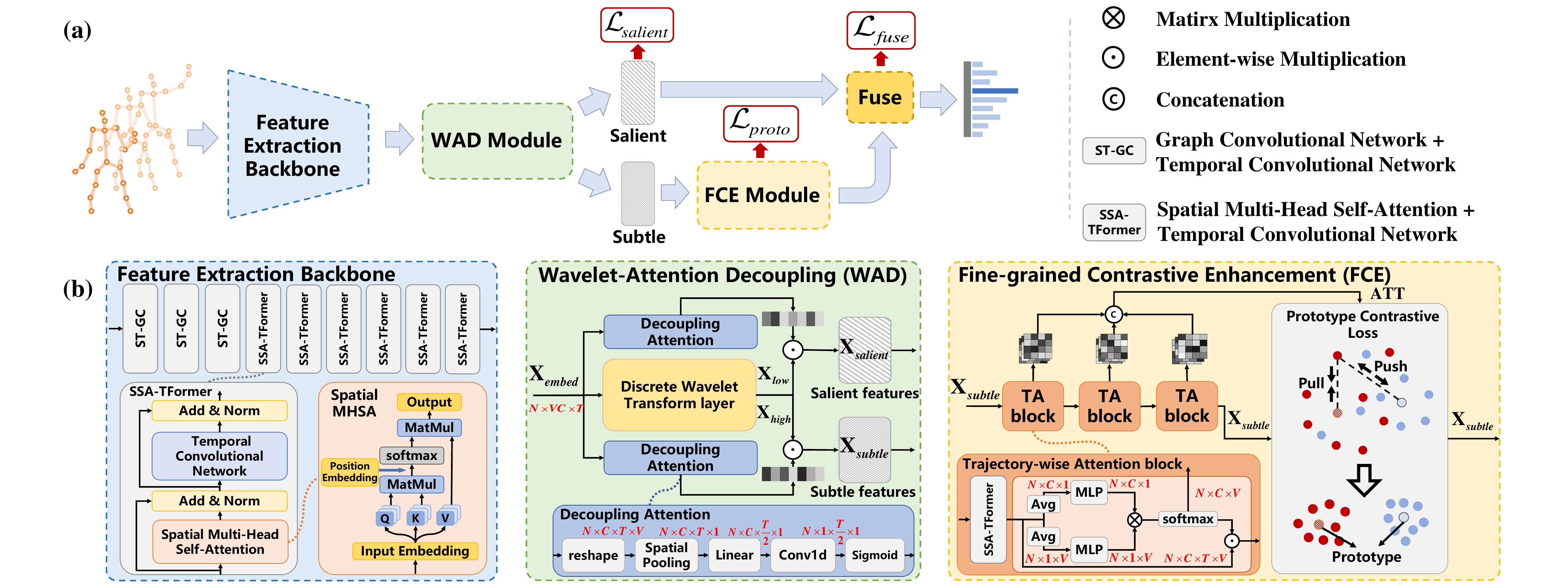}
	\vspace{-8mm}  
	\caption{{\bf (a)} Overview of the proposed WDCE-Net. {\bf (b)} Wavelet-Attention Decoupling (WAD) module maps the original features into the time-frequency domain and decouples salient and subtle motion features. Fine-grained Contrastive Enhancement (FCE) module enhances subtle features and amplifies the differences of confusing actions.}
	\vspace{-5mm}  
	\label{fig2}
\end{figure*} 

However, most existing approaches are not specifically designed for confusing actions, so it is difficult to distinguish similar fine-grained actions like “Reading” and “Writing”, as shown in Fig.\ref{fig1}. While there have been some efforts \cite{FOCUS,SFA,FRHead,Multi-Granular} to recognize confusing actions, research studies in this field remain limited. Geng et al. \cite{FOCUS} proposed a self-attention-enhanced graph neural network that fuses angle information to recognize confusing actions. Zhou et al. \cite{FRHead} utilized contrastive learning to discover and calibrate ambiguous samples to refine the representation of features. Nevertheless, these works are all limited to the spatiotemporal domain, easily leading to the network disregarding crucial distinctions among similar actions. Generally, the distinguishing characteristics of confusing actions are primarily reflected in motion trajectory features \cite{SGP}. Motion trajectories of joints, seen as 1D time series signals, are composed of multiple frequency components that signify distinct motion patterns. In the spatiotemporal domain, these motion patterns are frequently intertwined, making it possible for subtle patterns to be overshadowed by other salient ones during the feature extraction procedure. This provides a shortcut for the network to focus on easy-to-learned salient discrepancies rather than subtle differences that contain more discriminative information. To mine multi-granular motion patterns simultaneously, Chen et al. \cite{Multi-Granular} employed multiple temporal resolutions to represent coarse and fine-grained motion features. However, this approach still extracts features in the original temporal domain, thereby constraining its capability to magnify subtle motion details. Actually, the frequency components of signals can be readily decomposed in the frequency domain. This inspires us to perform a mapping of skeleton features into the frequency domain to decouple salient and subtle motion features effectively and with a greater emphasis on refining the latter.

To this end, we propose a Wavelet-Decoupling Contrastive Enhancement Network (WDCE-Net) for fine-grained action recognition (see Fig.\ref{fig1}). In this network, we design a Wavelet-Attention Decoupling (WAD) module to disentangle salient and subtle motion features by employing discrete wavelet transform \cite{WaveCNets,WaveDiff,Spectral} in the time-frequency domain. To achieve effective feature decoupling, we propose a decoupling attention mechanism that can adaptively recalibrate the temporal responses of decoupled features. Furthermore, we introduce a Fine-grained Contrastive Enhancement (FCE) module to enhance the distinctions in subtle motion features by effectively capturing the potential correlations among trajectory features in the time-frequency domain. To further augment the inter-class variation of subtle features, we utilize the prototype contrastive loss to regulate the learning of trajectory-wise attention, ensuring diverse attention for different action classes. In summary, the primary contributions of this paper are as follows: \\
\textbullet \quad We propose a Wavelet-Attention Decoupling (WAD) module that leverages the discrete wavelet transform to effectively decouple salient and subtle motion features within skeleton action sequences in the time-frequency domain. Additionally, it adaptively recalibrates their temporal responses via parametric decoupling attention.\\
\textbullet \quad We propose a Fine-grained Contrastive Enhancement (FCE) module to amplify the discrepancies in subtle motion features by capturing the correlation between trajectory features and using prototype contrastive loss.\\
\textbullet \quad We extensively experiment on NTU RGB+D and FineGYM datasets, comparing our methods with state-of-the-art models. Results demonstrate the significant improvement achieved by our methods in fine-grained action recognition.

\vspace{-1mm}
\section{METHODOLOGY}
\vspace{-1mm}
\label{sec:method}

The overview of WDCE-Net is depicted in Fig.\ref{fig2}(a). First, the feature extraction backbone maps the skeleton features into embedding space. Next, the WAD module effectively disentangles salient and subtle features in the time-frequency domain for subsequent utilization. Then, the subtle features will be fed into the FCE module to amplify the discrepancies in motion details by contrastive learning. Finally, the salient and subtle features are fused for action classification.

The feature extraction backbone we designed contains 3 ST-GC layers and 6 SSA-Tformer layers, as shown in Fig.\ref{fig2}(b). ST-GC is composed of GCN and TCN blocks, which are derived from \cite{STGCN}. SSA-Tformer consists of Spatial Multi-Head Self-Attention and TCN blocks. The ST-GC layers preliminary embed features, and the SSA-Tformer layers model coarse-grained spatial dependencies. After feature extraction, we obtain high-level embedding feature $\mathbf{X}_{{embed}} \in \mathbb{R}^{N \times C \times T \times V}$.

\vspace{-3mm}
\subsection{Wavelet-Attention Decoupling (WAD)}
\label{ssec:WAD}
{\bf Discrete Wavelet Transform layer.} WAD module takes the feature map $\mathbf{X}_{{embed}}$ after feature extraction as input, where $N$ is the batch size, $C$ is the channel dimensions, $T$ is the frame number in the skeleton sequences, and $V$ is the joint number. The trajectory of each joint can be regarded as a 1D signal so that the overall action features can be decoupled in the frequency domain by a 1D-DWT. To this end, we reshape  $\mathbf{X}_{{embed}} \in \mathbb{R}^{N \times C \times T \times V}$ into  $\mathbf{X}_{{embed}} \in \mathbb{R}^{N \times VC \times T }$. Here, $VC\!\times\! T$ represents the trajectory features of different joint channels, which contain rich differentiated information. DWT layer can be expressed as 
\begin{sequation}
	\setlength{\abovedisplayskip}{3pt} 
	\setlength{\belowdisplayskip}{3pt}
	\left\{\begin{array}{l}
		\mathbf{X}_{{low }}=\boldsymbol{D} \boldsymbol{W} \boldsymbol{T}_{\mathbf{L}}\left(\mathbf{X}_{{embed }}\right)=\mathbf{X}_{ {embed }} \mathbf{L} \\
		\mathbf{X}_{ {high }}=\boldsymbol{D} \boldsymbol{W} \boldsymbol{T}_{\mathbf{H}}\left(\mathbf{X}_{{embed }}\right)=\mathbf{X}_{ {embed }} \mathbf{H}
	\end{array}\right.,
	\label{DWT}
\end{sequation}
where $\mathbf{L}, \mathbf{H} \in \mathbb{R}^{N \times T \times \frac{T}{2}}$ are the matrix forms of Haar low and high-pass filters, respectively.  $\mathbf{X}_{low\!}\! \in \! \mathbb{R}^{N \times VC \times \frac{T}{2} }$ and $\mathbf{X}_{{high\!}}\! \in \! \mathbb{R}^{N \times VC \times \frac{T}{2} }$ are decomposed low and high-frequency components. After that, our model does not perform on the original skeleton sequence space but on the wavelet spectrum.\\
{\bf Decoupling Attention.} To achieve adaptive and comprehensive feature decoupling, we develop a Decoupling Attention block (see Fig.\ref{fig2}(b)) that utilizes parametric temporal-level attention to adaptively recalibrate the temporal responses of low and high-frequency components. Specifically, the spatial pooling layer averages the embedding feature in the spatial domain, and the linear layer maps the temporal dimension to half of the original. After the 1D convolutional layer and the Sigmoid function, the attention-weight vectors will be dot-multiplied with the low and high-frequency features to obtain the decoupled salient feature $\mathbf{X}_{ {salient}}$ and  $\mathbf{X}_{ {subtle}}$.

\vspace{-3mm}
\subsection{Fine-grained Contrastive Enhancement (FCE)}
\vspace{-1mm}
\label{ssec:FCE}
{\bf Trajectory-wise Attention.} To enhance subtle features, we design Trajectory-wise Attention blocks to amplify the discrepancies in trajectory features. The Trajectory-wise Attention selects discriminative trajectory features by capturing the correlation between the trajectory features of different channels of each joint. When the input feature is $\mathbf{X}_{ {subtle}} \in \mathbb{R}^{N \times C \times \frac{T}{2} \times V}$, trajectory-wise attention can be expressed as

%

\begin{sequation}
\setlength{\abovedisplayskip}{-5pt} 
\setlength{\belowdisplayskip}{3pt}
\resizebox{0.9\hsize}{!}{$\begin{aligned}& \mathbf{ATT}\!=\!\operatorname{softmax}\!\left(\mathbf{MLP}\!\left(\operatorname{Avg}\!\left(\mathbf{X}_{{subtle }}\!\right)\!\right)\! \otimes\! \mathbf{MLP}\!\left(\operatorname{Avg}\!\left(\mathbf{X}_{{subtle\! }}\right)\!\right)\!\right) \\ 
		& \mathbf{X}_{{subtle }}=\mathbf{X}_{{subtle}} \odot \mathbf{A T T} \end{aligned}$}
\label{TA}
\end{sequation}

Where $\mathbf{MLP}$ are fully connected layers. The two-way features output by MLPs are then transformed into a trajectory-wise attention map $\mathbf{ATT} \in \mathbb{R}^{N \times C \times V}$ after matrix multiplication and $\operatorname{softmax}$ function. By performing element-wise multiplication of $\mathbf{ATT}$ with $\mathbf{X}_{{subtle }}$, we are able to obtain the enhanced subtle features that contain rich discriminative information on motion trajectories. \\
{\bf Prototype contrastive loss.} To further enhance the subtle discrepancies, we design prototype contrastive loss to guide the learning process of Trajectory-wise Attention, enabling the capture of more discriminative trajectory correlations. To be specific, we maintain a feature prototype $P_k^{Feat}$ and an attention prototype $P_k^{ATT}$ , which are updated by the subtle features or attention maps of correctly classified action samples. To bring action samples closer to the prototypes of their own class and farther away from others, we design a prototype contrastive loss as
\begin{sequation}
	\setlength{\abovedisplayskip}{3pt} 
	\setlength{\belowdisplayskip}{3pt}
	\resizebox{0.9\hsize}{!}{$\begin{aligned}& \mathcal{L}_{{proto}}\!=\!\alpha \! \cdot\! \left(\!-\log \frac{e^{\operatorname{dis}\left(\mathbf{X}_{{subtle}}, P_k^{{Feat }}\right) / T}}{e^{\operatorname{dis}\left(\mathbf{X}_{{subtle}}, P_k^{{Feat }}\right) / T}+\sum_{l \neq k} e^{\operatorname{dis}\left(\mathbf{X}_{{subtle}}, P_l^{{Feat }}\right) / T}}\right) \\
			& +\beta \!\cdot\!\left(-\log \frac{e^{\operatorname{dis}\left(\mathbf{A T T}, P_k^{A T T}\right) / T}}{e^{\operatorname{dis}\left(\mathbf{A T T}, P_k^{A T T}\right) / T}+\sum_{l \neq k} e^{\operatorname{dis}\left(\mathbf{A T T}, P_l^{A T T}\right) / T}}\right)
		\end{aligned}$}
	\label{Lproto}
\end{sequation}
where $\operatorname{dis(\cdot)}$ represents the cosine distance, $\alpha$ and $\beta$ are hyper-parameters. The loss helps to highlight intra-class consistency and inter-class variability of trajectory features to make subtle features more discriminative.

\vspace{-3mm}
\subsection{Feature Fusion and Training Objective}
\label{ssec:Objective}

To leverage salient and subtle features simultaneously, we fuse by summing them together to obtain the fused feature $\mathbf{X}_{ {fuse}}$. Cross-entropy loss is applied to $\mathbf{X}_{ {fuse}}$ and $\mathbf{X}_{ {salient}}$, while prototype contrastive loss is used for $\mathbf{X}_{ {subtle}}$. Finally, the full learning objective function can be expressed as: 
\begin{equation}
	\setlength{\abovedisplayskip}{3pt} 
	\setlength{\belowdisplayskip}{3pt}
\mathcal{L}=\lambda_{ {fuse}} \cdot \mathcal{L}_{ {fuse}}+\lambda_{ {salient }} \cdot \mathcal{L}_{ {salient }}+\lambda_{ {proto }} \cdot \mathcal{L}_{ {proto }},
	\label{loss}
\end{equation}
where $\lambda_{ {fuse}}$, $ \lambda_{ {salient}}$ and $\lambda_{ {proto}}$ are hyper-parameters to trade off the three parts.

\section{EXPERIMENTS}
\label{sec:exper}

\subsection{Datasets and Settings}
\label{ssec:Datasets}
{\bf Datasets.} We evaluate our approach on coarse-grained dataset NTU RGB+D \cite{PLSTM}, and the fine-grained dataset FineGYM \cite{FineGYM}. For NTU RGB+D, we follow the data preprocessing in \cite{CTRGCN}. For FineGYM, we follow the method \cite{Pyskl} to extract the skeleton data. \\
{\bf Settings.} We implement our method with Pytorch framework and perform all experiments on one RTX 4090 GPU. We train our models using SGD with a momentum of 0.9 and weight decay of 0.0004. The batch size is set to 64 and the base learning rate is set to 0.1. The hyper-parameters in our methods are set as: $\alpha=0.9$, $\beta = 0.1$, $\lambda_{ {fuse}}=0.4$, $ \lambda_{ {salient}}=0.2$ and $\lambda_{ {proto}}=0.4$.  For NTU RGB+D, the learning rate decays with a factor of 0.1 at epoch 35, 55 and 75 for 130 epochs. For FineGYM, the learning rate decays with a factor of 0.1 at epoch 75, 115 and 155 for 200 epochs. 

\vspace{-3mm}
\subsection{Ablation Study}
\label{ssec:Ablation}
Tab.\ref{tab1} shows the results of our method under different components on the X-Sub benchmark of NTU RGB+D dataset. We only used the joint input modality in the experiments. We build the baseline with the backbone proposed in Section \ref{sec:method} and train it using the cross-entropy loss. In our experiments, “DWT” is the Discrete Wavelet Transform layer, “DA” is the Decoupling Attention, “TA” is the Trajectory-wise Attention block, and “PCL” is the prototype contrastive loss. It is evident that all these components contribute to the enhancement of the baseline's performance. Furthermore, when combined, they yield even better results. Besides, to verify the effect of decoupling features in the frequency domain, we replace the feature decoupling method from DWT with the channel split method which decouples features by splitting the channels in half in the spatiotemporal domain. The results show that decoupling subtle features in the frequency domain are more effective in identifying confusing actions.\\

\begin{table}
	\centering
	\caption{Comparisons of classification accuracies when applying the proposed components to the baseline.}
	\label{tab1}
	\vspace{-8pt}
	\setlength\tabcolsep{3pt} 
	\scalebox{0.7}[0.7]{
	\begin{tabular}{lccccc}

		\toprule
		\multirow{2}*{{\bf Method}}  & \multicolumn{2}{c}{{\bf WAD}} & \multicolumn{2}{c}{{\bf FCE}} & \multirow{2}*{{\bf Acc($\%$)}}\\
		\cmidrule(lr){2-3}\cmidrule(lr){4-5}
		& {\bf DWT} & {\bf DA} & {\bf TA} & {\bf PCL}\\
		\midrule
		Baseline & \small \XSolidBrush & \small\XSolidBrush & \small\XSolidBrush & \small\XSolidBrush & 88.8\\
		Baseline + DWT & \small\Checkmark & \small\XSolidBrush & \small\XSolidBrush & \small\XSolidBrush & 89.3\\
		Baseline + DWT + DA & \small\Checkmark & \small\Checkmark & \small\XSolidBrush & \small\XSolidBrush & 89.6\\ 
		Baseline + Channel Split + DA & \small\XSolidBrush & \small\Checkmark & \small\XSolidBrush & \small\XSolidBrush & 88.0\\
		Baseline + DWT + DA + PCL & \small\Checkmark & \small\Checkmark & \small\XSolidBrush & \small\Checkmark & 89.9\\
		Baseline + DWT + DA + TA & \small\Checkmark & \small\Checkmark & \small\Checkmark & \small\XSolidBrush & 89.8\\
		\midrule
		(Ours) Baseline + DWT +  DA + TA + PCL & \small\Checkmark & \small\Checkmark & \small\Checkmark & \small\Checkmark & {\bf 90.6}\\
		\bottomrule
		
	\end{tabular}
	}
\end{table}

\vspace{-7mm}
\subsection{Performance on fine-grained actions}
\label{ssec:Performance}
\vspace{-1mm}
\begin{figure}[t]
	\centering
	
	\includegraphics[width=\linewidth]{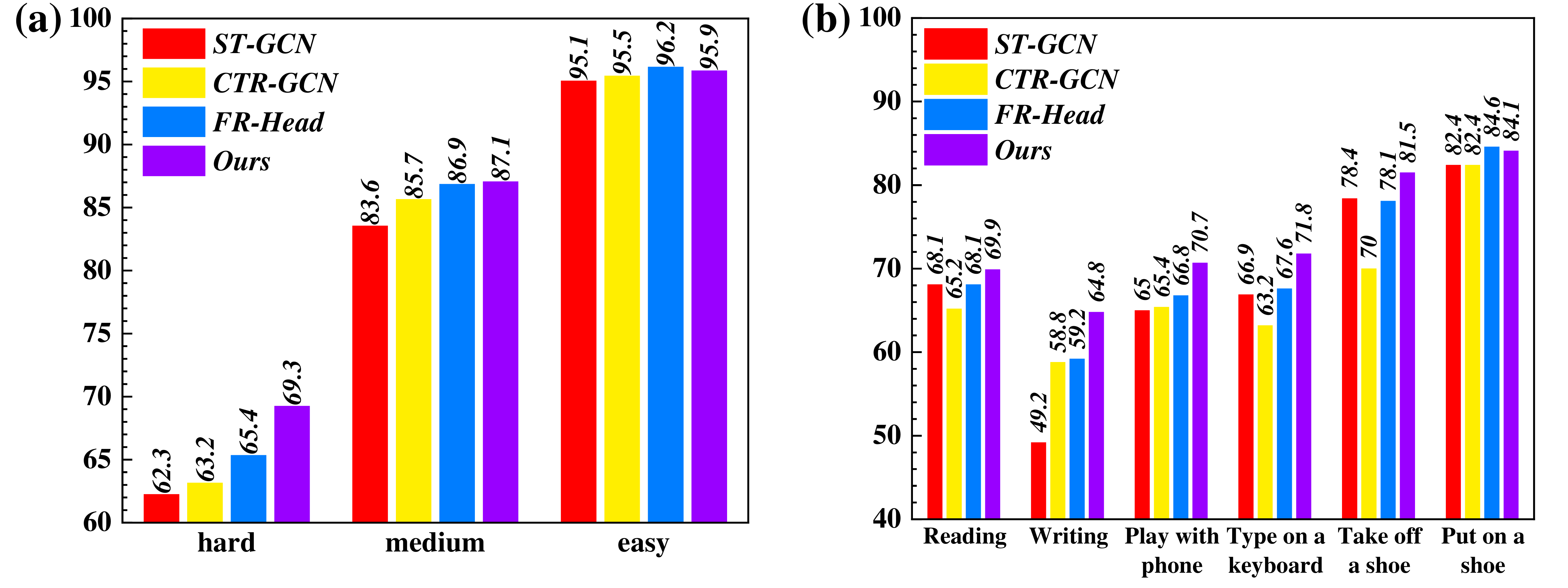}
	\vspace{-7mm}  
	\caption{Accuracy comparison of our method with ST-GCN, CTR-GCN and FR-Head. (a) Results on three sub-datasets. (b) Results on six easily confusing actions.}

	\vspace{-4mm}  
	\label{fig3}
\end{figure}

Same as \cite{FRHead}, we divide the NTU-RGB+D into three sub-datasets with different difficulty levels to verify the superiority of our method. According to the classification results of ST-GCN \cite{STGCN}, we select action samples whose accuracy is lower than 70$\%$ as a hard set, between 70$\%$ and 90$\%$ as a medium set, and over 90$\%$ as an easy set. Action samples in the hard set are usually confusing actions. As shown in Fig.\ref{fig3}(a), in the hard set and medium set, our method improves the classification accuracy compared with the other three methods \cite{STGCN,CTRGCN,FRHead}, and improves the most in the hard set. Fig.\ref{fig3}(b) shows that our method also achieves outstanding performance in identifying six easily confusing actions. The above results prove that our method has a great promotion in fine-grained action recognition. 

\begin{figure}[t]
	\centering
	
	\includegraphics[width=\linewidth]{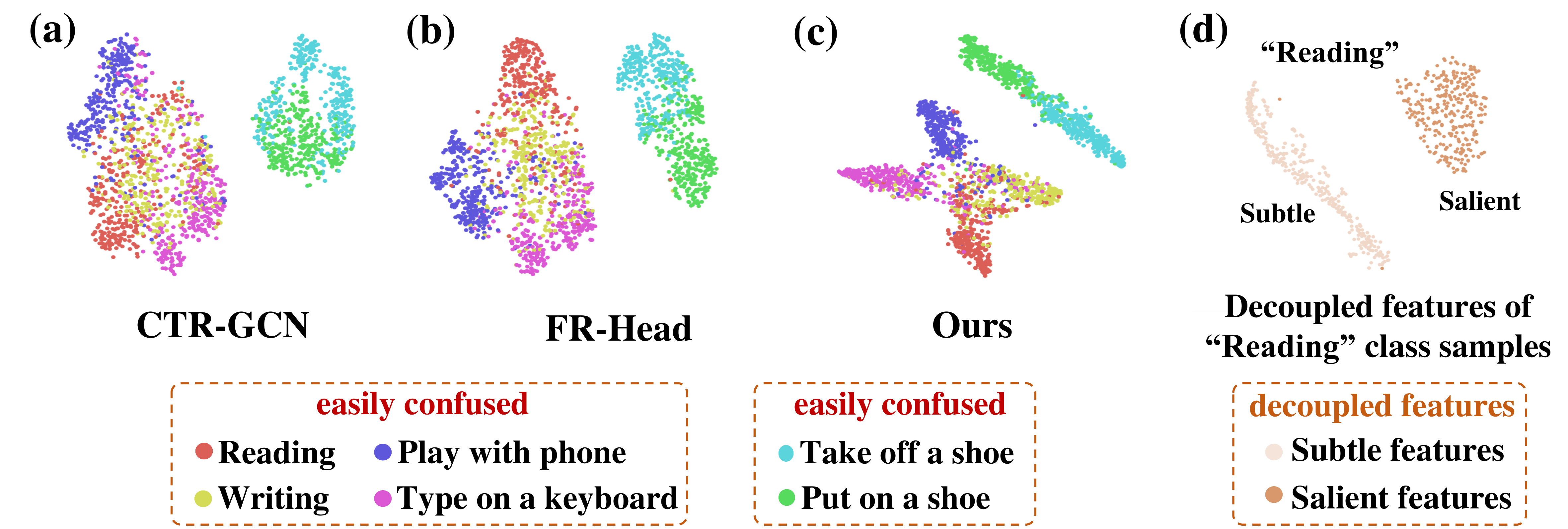}
	\vspace{-7mm}  
	\caption{Visualization of features by t-SNE. (a)$\sim$(c) Visualization results of CTR-GCN, FR-Head, and our method. (d) Feature decoupling results of “Reading” class samples.}

	\vspace{-4mm}  
	\label{fig4}
\end{figure}

Besides, we visualize the distribution of six confusing actions in the feature space using t-SNE as shown in Fig.\ref{fig4}. Note that our method (see Fig.\ref{fig4}(c)) can make confusing action features more compact than other methods (see Fig.\ref{fig4}(a) and Fig.\ref{fig4}(b)). At the same time, Fig.\ref{fig4}(d) proves that our method can disentangle the salient and subtle features of action samples effectively. Additionally, Fig.\ref{fig1} shows the clustering results of salient and subtle features of four confusing actions, which shows that salient features of confusing actions are difficult to distinguish, but subtle features contain more discriminative information.

\vspace{-3mm}
\subsection{Comparison with State-of-the-art Methods}
\label{ssec:Comparison}
In this section, we conduct a comparison with the state-of-the-art methods on NTU RGB+D \cite{PLSTM} and FineGYM \cite{FineGYM} datasets to demonstrate the competitive ability of our proposed method. Comparisons for each dataset are shown in Tab.\ref{tab2}. Following most previous approaches \cite{CTRGCN,FRHead}, we report the results after fusing the four modalities: joint, bone, joint motion, and bone motion. On NTU RGB+D, our method gets close to the state-of-the-art model HD-GCN \cite{HD-GCN} which fuses six modalities. But compared with the 4-streams version of HD-GCN, our model performs better. Besides, our model achieves state-of-the-art performance on the fine-grained dataset FineGYM, indicating better handling of fine-grained action recognition tasks.

\begin{table}
	\centering
	\caption{Comparisons of the top-1 accuracy($\%$) against state-of-the-art methods on the NTU-RGB+D and FineGYM.}
	\label{tab2}
	\vspace{-8pt}
	\setlength\tabcolsep{2pt} 
	\scalebox{0.7}[0.7]{
		\begin{tabular}{ccccc}
			\toprule
			{\bf Method}  & {\bf Publication} & {\bf \makecell[c]{NTU60\\X-Sub($\%$)}} & {\bf \makecell[c]{NTU60\\X-View($\%$)}} & {\bf FineGYM($\%$)} \\
			\midrule
			ST-GCN \cite{STGCN} & AAAI 2018 & 81.5 & 88.3 & 86.7 \\
			2s-AGCN \cite{2SAGCN} & CVPR 2019 & 88.5 & 95.1 & - \\
			MS-G3D \cite{MS-G3D} & CVPR 2020 & 91.5 & 96.2 & 92.0 \\
			CTR-GCN \cite{CTRGCN} & ICCV 2021 & 92.4 & 96.8 & 91.9 \\
			EfficientGCN-B4 \cite{EfficientGCN-B4} & TPAMI 2022 & 91.7 & 95.7 & - \\
			InfoGCN (6-streams) \cite{InfoGCN} & CVPR 2022 & 93.0 & 97.1 & 92.0 \\
			FR-Head (4-streams) \cite{FRHead} & CVPR 2023 & 92.8 & 96.8 & - \\
			HD-GCN (4-streams) \cite{HD-GCN} & ICCV 2023 & 93.0 & 97.0 & - \\
			HD-GCN (6-streams) \cite{HD-GCN} & ICCV 2023 & {\bf 93.4} & 97.2 & - \\
			\midrule
			Ours (4-streams) &  & 93.0 & {\bf 97.2} & {\bf 93.9} \\
			\bottomrule
			
		\end{tabular}
	}
\end{table}

\section{CONCLUSION}
\label{sec:conclu}
This paper presents a Wavelet-Decoupling Contrastive Enhancement Network (WDCE-Net) for fine-grained skeleton-based action recognition. WDCE-Net leverages discrete wavelet transform recalibrated by decoupling attention mechanism to decouple salient and subtle motion features. Then WDCE-Net utilizes prototype contrastive loss to guide the learning of Trajectory-wise Attention to mine discriminative patterns of subtle motion features. Extensive experiments show that the proposed WDCE-Net has a large performance advantage in distinguishing confusing actions.

%


\vfill\pagebreak

\bibliographystyle{IEEEbib}
\bibliography{refs}

\begin{thebibliography}{10}

\bibitem{PLSTM}
Amir Shahroudy, Jun Liu, Tian-Tsong Ng, and Gang Wang,
\newblock ``Ntu rgb+ d: A large scale dataset for 3d human activity analysis,''
\newblock in {\em Proceedings of the IEEE conference on computer vision and
  pattern recognition}, 2016, pp. 1010--1019.

\bibitem{HCN}
Chao Li, Qiaoyong Zhong, Di~Xie, and Shiliang Pu,
\newblock ``Co-occurrence feature learning from skeleton data for action
  recognition and detection with hierarchical aggregation,''
\newblock {\em arXiv preprint arXiv:1804.06055}, 2018.

\bibitem{Body}
Qianshuo Hu, Hong Liu, Hua-Qiu Wang, and Mengyuan Liu,
\newblock ``Body prior guided graph convolutional neural network for
  skeleton-based action recognition,''
\newblock in {\em Proceedings of the IEEE International Conference on
  Acoustics, Speech and Signal Processing}. IEEE, 2023, pp. 1--5.

\bibitem{STGCN}
Sijie Yan, Yuanjun Xiong, and Dahua Lin,
\newblock ``Spatial temporal graph convolutional networks for skeleton-based
  action recognition,''
\newblock in {\em Proceedings of the AAAI conference on artificial
  intelligence}, 2018, vol.~32.

\bibitem{2SAGCN}
Lei Shi, Yifan Zhang, Jian Cheng, and Hanqing Lu,
\newblock ``Two-stream adaptive graph convolutional networks for skeleton-based
  action recognition,''
\newblock in {\em Proceedings of the IEEE/CVF conference on computer vision and
  pattern recognition}, 2019, pp. 12026--12035.

\bibitem{CTRGCN}
Yuxin Chen, Ziqi Zhang, Chunfeng Yuan, Bing Li, Ying Deng, and Weiming Hu,
\newblock ``Channel-wise topology refinement graph convolution for
  skeleton-based action recognition,''
\newblock in {\em Proceedings of the IEEE/CVF international conference on
  computer vision}, 2021, pp. 13359--13368.

\bibitem{MS-G3D}
Ziyu Liu, Hongwen Zhang, Zhenghao Chen, Zhiyong Wang, and Wanli Ouyang,
\newblock ``Disentangling and unifying graph convolutions for skeleton-based
  action recognition,''
\newblock in {\em Proceedings of the IEEE/CVF conference on computer vision and
  pattern recognition}, 2020, pp. 143--152.

\bibitem{EfficientGCN-B4}
Yi-Fan Song, Zhang Zhang, Caifeng Shan, and Liang Wang,
\newblock ``Constructing stronger and faster baselines for skeleton-based
  action recognition,''
\newblock {\em IEEE transactions on pattern analysis and machine intelligence},
  vol. 45, no. 2, pp. 1474--1488, 2022.

\bibitem{SGP}
Yuxin Chen, Gaoqun Ma, Chunfeng Yuan, Bing Li, Hui Zhang, Fangshi Wang, and
  Weiming Hu,
\newblock ``Graph convolutional network with structure pooling and joint-wise
  channel attention for action recognition,''
\newblock {\em Pattern Recognition}, vol. 103, pp. 107321, 2020.

\bibitem{InfoGCN}
Hyung-gun Chi, Myoung~Hoon Ha, Seunggeun Chi, Sang~Wan Lee, Qixing Huang, and
  Karthik Ramani,
\newblock ``Infogcn: Representation learning for human skeleton-based action
  recognition,''
\newblock in {\em Proceedings of the IEEE/CVF Conference on Computer Vision and
  Pattern Recognition}, 2022, pp. 20186--20196.

\bibitem{HD-GCN}
Jungho Lee, Minhyeok Lee, Dogyoon Lee, and Sangyoun Lee,
\newblock ``Hierarchically decomposed graph convolutional networks for
  skeleton-based action recognition,''
\newblock {\em arXiv preprint arXiv:2208.10741}, 2022.

\bibitem{STTR}
Chiara Plizzari, Marco Cannici, and Matteo Matteucci,
\newblock ``Skeleton-based action recognition via spatial and temporal
  transformer networks,''
\newblock {\em Computer Vision and Image Understanding}, vol. 208, pp. 103219,
  2021.

\bibitem{FOCUS}
Pei Geng, Xuequan Lu, Chunyu Hu, Hong Liu, and Lei Lyu,
\newblock ``Focusing fine-grained action by self-attention-enhanced graph
  neural networks with contrastive learning,''
\newblock {\em IEEE Transactions on Circuits and Systems for Video Technology},
  2023.

\bibitem{SFA}
Kaiyuan Liu, Yunheng Li, Yuanfeng Xu, Shuai Liu, and Shenglan Liu,
\newblock ``Spatial focus attention for fine-grained skeleton-based action
  tasks,''
\newblock {\em IEEE Signal Processing Letters}, vol. 29, pp. 1883--1887, 2022.

\bibitem{FRHead}
Huanyu Zhou, Qingjie Liu, and Yunhong Wang,
\newblock ``Learning discriminative representations for skeleton based action
  recognition,''
\newblock in {\em Proceedings of the IEEE/CVF Conference on Computer Vision and
  Pattern Recognition}, 2023, pp. 10608--10617.

\bibitem{Multi-Granular}
Tailin Chen, Desen Zhou, Jian Wang, Shidong Wang, Yu~Guan, Xuming He, and Errui
  Ding,
\newblock ``Learning multi-granular spatio-temporal graph network for
  skeleton-based action recognition,''
\newblock in {\em Proceedings of the 29th ACM international conference on
  multimedia}, 2021, pp. 4334--4342.

\bibitem{WaveCNets}
Qiufu Li, Linlin Shen, Sheng Guo, and Zhihui Lai,
\newblock ``Wavelet integrated cnns for noise-robust image classification,''
\newblock in {\em Proceedings of the IEEE/CVF Conference on Computer Vision and
  Pattern Recognition}, 2020, pp. 7245--7254.

\bibitem{WaveDiff}
Hao Phung, Quan Dao, and Anh Tran,
\newblock ``Wavelet diffusion models are fast and scalable image generators,''
\newblock in {\em Proceedings of the IEEE/CVF Conference on Computer Vision and
  Pattern Recognition}, 2023, pp. 10199--10208.

\bibitem{Spectral}
Tommi Kerola, Nakamasa Inoue, and Koichi Shinoda,
\newblock ``Spectral graph skeletons for 3d action recognition,''
\newblock in {\em Asian conference on computer vision}. Springer, 2014, pp.
  417--432.

\bibitem{FineGYM}
Dian Shao, Yue Zhao, Bo~Dai, and Dahua Lin,
\newblock ``Finegym: A hierarchical video dataset for fine-grained action
  understanding,''
\newblock in {\em Proceedings of the IEEE/CVF conference on computer vision and
  pattern recognition}, 2020, pp. 2616--2625.

\bibitem{Pyskl}
Haodong Duan, Jiaqi Wang, Kai Chen, and Dahua Lin,
\newblock ``Pyskl: Towards good practices for skeleton action recognition,''
\newblock in {\em Proceedings of the 30th ACM International Conference on
  Multimedia}, 2022, pp. 7351--7354.

\end{thebibliography}
\end{document}